\documentclass[sigconf]{acmart}
\AtBeginDocument{%
  }

\copyrightyear{2025}
\acmYear{2025}
\setcopyright{rightsretained}

\acmConference[WWW Companion '25]{Companion Proceedings of the ACM Web Conference 2025}{April 28-May 2, 2025}{Sydney, NSW, Australia}
\acmBooktitle{Companion Proceedings of the ACM Web Conference 2025 (WWW Companion '25), April 28-May 2, 2025, Sydney, NSW, Australia}
\acmDOI{10.1145/3701716.3715523}
\acmISBN{979-8-4007-1331-6/25/04}

\makeatletter
\def\@ACM@checkaffil{% Only warnings
    \if@ACM@instpresent\else
    \ClassWarningNoLine{\@classname}{No institution present for an affiliation}%
    \fi
    \if@ACM@citypresent\else
    \ClassWarningNoLine{\@classname}{No city present for an affiliation}%
    \fi
    \if@ACM@countrypresent\else
        \ClassWarningNoLine{\@classname}{No country present for an affiliation}%
    \fi
}
\makeatother

\makeatletter
\gdef\@copyrightpermission{
  \begin{minipage}{0.2\columnwidth}
   \includegraphics[width=0.90\textwidth]{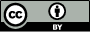}
  \end{minipage}\hfill
  \begin{minipage}{0.8\columnwidth}
   \href{https://creativecommons.org/licenses/by/4.0/}{This work is licensed under a Creative Commons Attribution International 4.0 License.}
  \end{minipage}
  \vspace{5pt}
}
\makeatother

\usepackage{graphicx}
\usepackage{multirow}
\usepackage{amsmath}
\usepackage{relsize}
\usepackage{bbold}
\usepackage{caption} 
\usepackage{epsfig}
\usepackage{epstopdf}
\usepackage{algorithmic}
\usepackage{algorithm}
\usepackage{hyperref}
\usepackage{verbatim}
\usepackage{multirow}
\usepackage{pifont}% http://ctan.org/pkg/pifont
\usepackage{graphics}
\usepackage{adjustbox}
\usepackage{boldline} 
\usepackage{float}
\usepackage{pdflscape}
\usepackage{afterpage}
\usepackage{capt-of}% or use the larger `caption` package
\usepackage{subfig}
\usepackage{bbm}
\usepackage[toc,page]{appendix}
\usepackage{balance}
\usepackage{url}
\makeatletter

\graphicspath{ {./images/} }

\usepackage[rightcaption]{sidecap}
\usepackage{wrapfig}

\usepackage{booktabs}
\usepackage{placeins}
\usepackage[labelfont=bf]{caption}
\usepackage[normalem]{ulem}
\usepackage{color}
\usepackage{soul}
\usepackage{tikz} % only to get \foreach
\usepackage{hyperref}
\usepackage{paralist}
\usepackage[inline]{enumitem}

\copyrightyear{2025}
\acmYear{2025}
\setcctype{by}
\acmConference[WWW Companion '25]{Companion Proceedings of the ACM Web Conference 2025}{April 28-May 2, 2025}{Sydney, NSW, Australia}
\acmBooktitle{Companion Proceedings of the ACM Web Conference 2025 (WWW Companion '25), April 28-May 2, 2025, Sydney, NSW, Australia}
\acmDOI{10.1145/3701716.3715523}
\acmISBN{979-8-4007-1331-6/25/04}

\begin{document}

\title{Uncertainty-Aware Fusion: An Ensemble Framework for Mitigating Hallucinations in Large Language Models}

\author{Prasenjit Dey}
\affiliation{%
  \institution{Amazon}
%  \city{Central Machine Learning Team}
  }
\email{prasendx@amazon.com}

\author{Srujana Merugu}
\affiliation{%
  \institution{Amazon}
%  \city{Central Machine Learning Team}
  }
\email{smerugu@amazon.com}

\author{Sivaramakrishnan Kaveri}
\affiliation{%
  \institution{Amazon}
%  \city{Central Machine Learning Team}
  }
\email{kavers@amazon.com}


\begin{abstract}
Large Language Models (LLMs) are known to hallucinate and generate non-factual outputs which can undermine user trust. Traditional methods to directly mitigate hallucinations, such as representation editing and contrastive decoding, often require additional training data and involve high implementation complexity. While ensemble-based approaches harness multiple LLMs to tap into the "wisdom of crowds",  these methods overlook uncertainties in individual model responses. Recent studies reveal that uncertainty estimation can enable LLMs to self-assess the likelihood of generating hallucinations. In this work, we focus on factoid question answering (QA) and observe that LLMs accuracy and self-assessment capabilities vary widely with  different models excelling in different scenarios. Leveraging this insight, we propose Uncertainty-Aware Fusion (UAF), an ensemble framework to reduces hallucinations by strategically combining multiple LLM based on their accuracy and self-assessment  abilities. Empirical results on several public benchmark datasets show that UAF outperforms state-of-the-art hallucination mitigation methods by $8\%$ in factual accuracy, while either narrowing or surpassing the performance gap with GPT-4.

\end{abstract}





\maketitle

%% ========================================================
\newcommand\R{\mathbb{R}}
\newcommand\U{\mathbb{U}}
\newcommand\eS{\mathbb{S}} % why did we do this
\newcommand\I{\mathbb{I}}
\newcommand\1{\mathbbm{1}}

%% ========================================================
%% Some useful commands. Note the dangerous \r command!
%% ========================================================
\newcommand{\bd}{\boldsymbol}
\newcommand{\mf}{\mathbf}

\renewcommand{\a}{\mathbf{a}}
\renewcommand{\c}{\mathbf{c}}
\renewcommand{\r}{\mathbf{r}} % since \r was already defined in altex
\renewcommand{\u}{\mathbf{u}}
\renewcommand{\v}{\mathbf{v}}
\newcommand{\s}{\mathbf{s}}
\newcommand{\w}{\mathbf{w}}
\newcommand{\x}{\mathbf{x}}
\newcommand{\y}{\mathbf{y}}
\newcommand{\z}{\mathbf{z}} 
\newcommand{\p}{\mathbf{p}}
\newcommand{\myb}{\mathbf{b}}
\newcommand{\q}{\mathbf{q}}

%----------------------------------------
%      THE SETS
%----------------------------------------
\newcommand{\cC}{{\cal C}}
\newcommand{\cD}{{\cal D}}
\newcommand{\cE}{{\cal E}}
\newcommand{\cU}{{\cal U}}
\newcommand{\cL}{{\cal L}}
\newcommand{\cM}{{\cal M}}
\newcommand{\cP}{{\cal P}}
\newcommand{\cQ}{{\cal Q}}
\newcommand{\cS}{{\cal S}}
\newcommand{\cT}{{\cal T}}
\newcommand{\cX}{{\cal X}}
\newcommand{\cY}{{\cal Y}}
\newcommand{\cZ}{{\cal Z}}
\newcommand{\cF}{{\cal F}}
\newcommand{\cB}{{\cal B}}
\newcommand{\cA}{{\cal A}}

%--------------------------------------------
%         THE VARIABLES & APPROXIMATES
%--------------------------------------------
\newcommand{\bM}{\mathbf{M}}
\newcommand{\bW}{\mathbf{W}}
\newcommand{\bA}{\mathbf{A}}
\newcommand{\bB}{\mathbf{B}}
\newcommand{\bR}{\mathbf{R}}
\newcommand{\bC}{\mathbf{C}}
\newcommand{\bE}{\mathbf{E}}
\newcommand{\bZ}{\mathbf{Z}}
\newcommand{\bG}{\mathbf{G}}

\newcommand{\hP}{\hat{P}}
\newcommand{\hZ}{\hat{Z}}
\newcommand{\hM}{\hat{M}}
\newcommand{\hA}{\hat{A}}
\newcommand{\hC}{\hat{C}}
\newcommand{\hU}{\hat{U}}
\newcommand{\hV}{\hat{V}}
\newcommand{\hX}{\hat{X}}
\newcommand{\hY}{\hat{Y}}
\newcommand{\hG}{\hat{G}}
\newcommand{\hS}{\hat{S}}

\newcommand{\Hl}{\hat{l}}
\newcommand{\hz}{\hat{z}}
\newcommand{\hu}{\hat{u}}
\newcommand{\hv}{\hat{v}}
\newcommand{\hx}{\hat{x}}
\newcommand{\hy}{\hat{y}}
\newcommand{\hp}{\hat{p}}
\newcommand{\Hm}{\hat{m}}

%---------MISC-----------------------------
\newcommand{\pr}{\mathbf{P}}
\newcommand{\E}{\mathbf{E}}

\newcommand{\xx}{\boldsymbol{\chi}}
\newcommand{\Th}{\boldsymbol{\theta}}
\newcommand{\Eta}{\boldsymbol{\eta}}

\newcommand{\norm}[2]{\ensuremath \|#1\|_{#2}}
\newcommand{\myref}[1]{(\ref{#1})}
\newcommand{\del}[2]{\frac{\partial #1}{\partial #2}}

\newcommand{\proofsketch}{\noindent{\itshape Proof Sketch:}\hspace*{1em}}
%\newcommand{\qed}{\nolinebreak[1]~~~\hspace*{\fill} \rule{5pt}{5pt}\vspace*{\parskip}\vspace*{1ex}}

%%%%%%%%%%%%%%%%%%%%%%%%%%%% notations %%%%%%%%%%%%%%%%%%%%%%%%%%%%%%%

% Count of LLMs ---> N
% subset of LLMs ---> K
% dataset ---> D
% D_{val} and D_{test} for validation and testset
% input output space \mathcal{X}, \mathcal{Y}$
% datapoint--->  x,y
% count of datapoints ---> n
% count of validation datapoints --->n_val
% Accuracy of ith llm ---> Acc_{i}
% uncertainty_auroc of ith llm  ---> Unc\_auroc_{i}
% combined score ---> C\_score
% individual jth model responses/output for ith datapoint ---> \hat{y_i^j}
% individual jth model uncertainty output for ith datapoint ---> u_i^j
% final output response for ith datapoint \hat{y_i}
% uncertainty of ith LLM ---> u_i

\begin{figure}
\centering
\includegraphics[width=.3\textwidth]{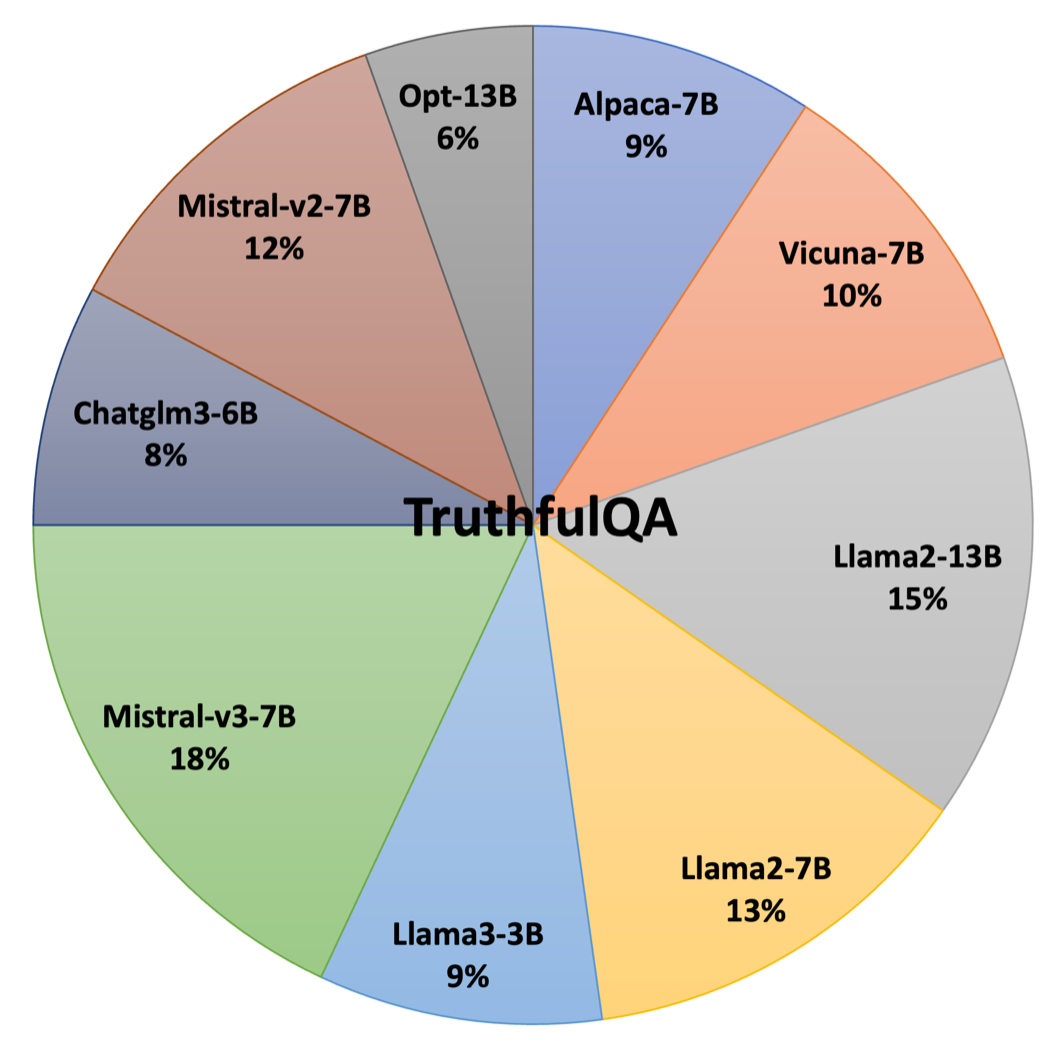}
\caption{Fraction of examples where each LLM generates the most confident correct answer. Optimal LLM ( high accuracy and confidence) varies across examples.}
\label{fig:pie}
\end{figure}

\section{Introduction}
\label{sec:intro}
Large language models (LLMs) have yielded remarkable performance boosts across diverse natural language processing (NLP) tasks \cite{gpt4}. However,  their tendency to 'hallucinate', i.e., generating outputs not grounded in factual training data\cite{hal_survey}, remains a critical limitation, 
impeding deployment, particularly in high-stakes applications where factual accuracy is paramount.\\
\textbf{Related work:} 
Existing work on hallucination mitigation can be broadly categorized into three groups.
The first group of techniques tries to enhance truthfulness at inference time using contrastive decoding or representation editing. Representation editing methods like ITI \cite{iti} and TruthX \cite{truthx} identify a truthful direction using human-labeled data and adjust model activations accordingly. Contrastive decoding approaches such as SH2 \cite{sh2} and DoLa \cite{dola} modify output probabilities by comparing distributions between different models or layers. However, these methods often face challenges with implementation complexity, reliance on extensive labeled data, and limited generalization across domains. \\
The second group employs ensemble learning techniques to improve model performance. Methods such as LLM-Blender \cite{llmblender} learn to combine diverse candidate responses from multiple LLMs, while MLM \cite{mlm} selects the response most similar to the input. Another approach exemplified by Consistent \cite{wang2022self}, generates multiple candidate responses and uses majority voting for the final output. However, these approaches often overlook the inherent uncertainties in candidate responses, potentially limiting their effectiveness.\\
The third group of methods is based on the observation that LLMs often have an internal sense of truthfulness, even when producing false statements  \cite{kadavath2022language,eigen}. This insight is exploited in various approaches such as logit probability-based methods \cite{selgen,gales_unc}, consistency-based techniques \cite{semantic_unc,selfckgpt}, probing internal representations \cite{eigen,haloscope}, and prompting-based strategies \cite{kadavath2022language} to obtain uncertainty scores without additional training or model editing, offering promising avenues for hallucination detection and mitigation.\\
\textbf{Contributions: } In this work, we focus on mitigating LLM hallucinations for factoid question answering (QA). First, we observe that the proliferation of open-source LLMs, each with distinct strengths and weaknesses, precludes any single model from consistently outperforming others in factual accuracy and self-assessment. Figure \ref{fig:pie} illustrates this phenomenon for the TruthfulQA \cite{truthfulqa} dataset, showing the distribution of examples where each LLM produces the most confident correct response. Despite Mistralv3-7B achieving the highest overall accuracy, it leads in confidence for only 18\% of examples, indicating that the rest are better handled by other models. This diversity in model performance across examples presents an opportunity to leverage the complementary strengths of multiple LLMs to reduce hallucinations and enhance overall accuracy. Our key contributions are summarized below:\\
%\begin{enumerate}
\noindent 1. We demonstrate, through extensive experiments with six LLMs across multiple benchmarks, that the relative factual accuracy and self-assessment capabilities of LLMs  vary significantly across examples, with no single dominant model. 

\noindent 2. We propose Uncertainty-Aware Fusion (UAF), an ensemble framework that strategically combines multiple LLMs to achieve superior performance. UAF consists of two key modules: (i) SELECTOR, which chooses the top $K$ LLMs from the entire pool based on accuracy and hallucination detection abilities, and (ii) FUSER, which merges the  outputs of these $K$ LLMs to generate the final result.

\noindent 3. Experiments on benchmarks such as  TruthfulQA \cite{truthfulqa}, TriviaQA \cite{triviaqa}, and FACTOR \cite{factor} datasets reveal that our ensembling technique significantly enhances factual accuracy with  UAF outperforming SOTA baselines by 8\% in accuracy, while narrowing the performance gap with, or even surpassing, GPT-4 \cite{gpt4}. We also present ablation studies examining the impact of uncertainty measure, model selection criteria, and ensemble size on overall performance.

\begin{figure}
\centering
\includegraphics[width=.5\textwidth]{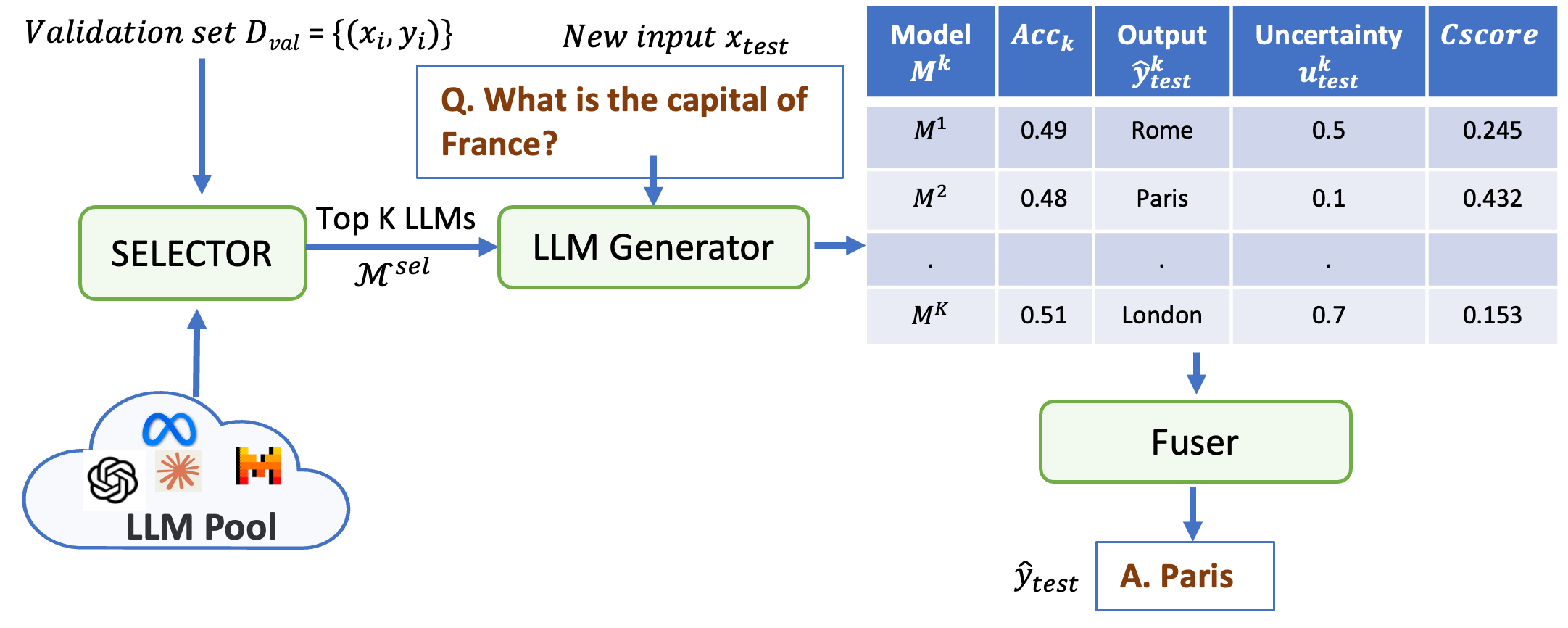}
\caption{Overview of our UAF architecture}
\label{fig:fuser}
\end{figure}

\vspace{-1pt}
\section{Proposed Method}
\textbf{Problem Statement: } Let $\mathcal{X}$ and $\mathcal{Y}$ denote the input and output spaces, respectively, and $D = \{(x_i, y_i)\}_{i=1}^n$ the dataset, where $x_i \in \mathcal{X}$ and $y_i \in \mathcal{Y}$ are the $i^{th}$ question-answer pair. For each $x_i$, the goal is to generate a response $\hat{y}_i$ that maximizes the overall accuracy. The goal is to achieve this using a pool  of $N$ pretrained foundational LLMs without additional training or fine-tuning.

\begin{comment}

Let $\mathcal{X}$ and $\mathcal{Y}$ denote input and output spaces respectively and  $\mathcal{D}= \{(x_i,y_i)\}_{i=1}^n$ denote Dataset, where $x_i \in \mathcal{X}$ and $y_i \in \mathcal{Y}$ is the $i^{th}$ question-answer(QA) pair. For each $x_i$ we want to generate a response $\hat{y}_i$  such that overall accuracy denoted by $\frac{1}{n}\sum_{i=1}^n\mathbf{I}(y_i=\hat{y}_i)$ is maximized.  We want to maximize this using ensemble of $N$ LLMs without any additional training or fine-tuning.
\end{comment}
\renewcommand{\algorithmicrequire}{\textbf{Input:}}
\renewcommand{\algorithmicensure}{\textbf{Output:}}

\begin{algorithm}[t]
    \caption{Components of UAF - SELECTOR and FUSER}
    \label{alg:selector}
    
    \begin{algorithmic}
    \REQUIRE $D_{val}$, Pool of LLMs $\mathcal{M}= \{M^j\}_{j=1}^{N}$, Uncertainty function $U_f(.,.,.)$, Ensemble size $K$, Test data point $x_{test}$
    \ENSURE Test data response $\hat{y}_{test}$
    \newline

    \STATE \textbf{procedure }SELECTOR ($D_{val}, \mathcal{M}, U_f, K$)
    \FOR{$M^j \in \mathcal{M}$}
    %\STATE $L_j, U_j = \emptyset,\emptyset$
    \FOR{$(x_i, y_i) \in D_{val}$}
    \STATE $\hat{y}_i^j = M^j(x_i)$   \quad;\quad  $s_i^j = \mathbf{1}(\hat{y}_i^j == y_i)$
    \STATE $u_i^j = U_f(M^j, x_i, \hat{y}_i^j)$
    %\STATE $L_j \leftarrow L_j \cup IsCorrect_i^j$   \quad;\quad    $U_j \leftarrow U_j \cup u_{val_i}^j$
    \ENDFOR

    \STATE $Acc_j = \frac{1}{|D_{val}|}\sum_i s_i^j$  ;\quad $SAH_j = ROC\_AUC\_score(\{s_i^j,u_i^j\}_i)$
    \STATE $Cscore_j = Acc_j \times SAH_j$
    \ENDFOR
    \STATE \textbf{return:} $\mathcal{M}^{sel} =  TopK (\{Cscore_j\}_{j=1}^N)$,  \hfill\COMMENT{//Top $K$ LLMs}
    \STATE \quad  \quad      $Acc^{sel} = \{Acc_j| j \in \mathcal{M}^{sel}\}$ \hfill\COMMENT{//Accuracy of selected $K$ LLMs}
    %\STATE $j_1, j_2, \dots, j_K = \arg\max_{j \in \{1, 2, \dots, N\}} Cscore_j$ 
    %\ENSURE $j_1, j_2, \dots, j_K$    \hfill\COMMENT{//Indices of Top K llms}
    \newline
\STATE \textbf{procedure } FUSER ($x_{test}, \mathcal{M}^{sel}, Acc^{sel}, U_f, K$)
\FOR{$M^k \in \mathcal{M}^{sel}$}
\STATE $\hat{y}_{test}^k = M^k(x_{test})$
\STATE $u_{test}^k = U_f(M^k, x_{test}, \hat{y}_{test}^k)$
\ENDFOR
\STATE $\hat{y}_{test} = \hat{y}_{test}^{k^*} \text{, where } k^* = argmax_{k \in \{1,\dots K\}} Acc_k \times (1-u_{test}^k)$
\STATE \textbf{return:} $\hat{y}_{test}$

\end{algorithmic}
\end{algorithm}

\subsection{Uncertainty Aware Fusion (UAF)}\label{sec:uaf}
Figure \ref{fig:fuser} provides an overview of our UAF framework. At a high level, UAF consists of two modules: SELECTOR and FUSER. Given a specific task, the SELECTOR selects the top $K$ LLMs from a pool of $N$ LLMs based on performance metric. FUSER then combines the outputs of these $K$ LLMs to produce the final response.

\subsubsection{SELECTOR}
Given a pool of $N$ LLMs denoted by $\mathcal{M}$, SELECTOR selects $K$ LLMs (where $K<N$) to optimize computational efficiency and enhance overall factual accuracy by pruning underperforming LLMs. Selection is based on two criteria: (1) task-specific accuracy and (2) self-assessment of hallucinations based on an given uncertainty measure. Given a specified uncertainty measure $U_f(\cdot)$, and  a validation set $D_{val}$, we prompt each LLM with input $x_i$,   obtaining response  $\hat{y}_i^j$  and corresponding uncertainty score $u_i^j$ from the $j^{th}$ LLM $M^j$.  We compute the accuracy $Acc_j$ of $M^j$ as the fraction of correct responses. We also measure the LLMs ability for  self-assessment of hallucinations $SAH_j$ as the area under the ROC curve for the binary classification of truthful vs. hallucinatory responses using uncertainty scores. We then compute a combined score $Cscore_j = Acc_j \times SAH_j$ for each LLM. The top K models with the highest combined scores are selected greedily, where K is a   hyperparameter tuned for specific tasks.

\begin{comment}
To achieve this we sample $10\%$ of examples from $D$ and denote it as  $D_{val}$. We use the rest denoted by $D_{test}$ for evaluation. For each $(xval_i,yval_i)\in D_{val}$ we prompt each of the $N$ LLMs with $xval_i$. Let $\hat{yval}_i^j$ denote the response and $u_i^j$ its corresponding uncertainty score computed with a particular uncertainty method for $j^{th}$ LLM. Accuracy of $j^{th}$ LLM, denoted by $Acc_j$ is defined as percentage of correct responses. We also measure area under the receiver operator characteristic curve  $Unc\_auroc_j$ of $j^{th}$ LLM, by measuring the performance of classifying its own correct(truthful) from incorrect(hallucinatory) responses by varying the thresholds on the set of uncertainty scores $\{u_i^j\}_{i=1}^{nval}$. For each $j^{th}$ LLM we compute a combined score $Cscore_j = Acc_j*Unc\_auroc_j$. We use greedy method to select $K$ LLMs based on top $K$ highest $Cscore$. Here $K$ is the hyperparameter which we tune using $D_{val}$. Algorithm \ref{alg:selector} presents the pseudo-code for this method.
\end{comment}

\subsubsection{FUSER}
Given the selected ensemble of $K$ models $\mathcal{M}^{sel}$, 
%with respective accuracies $\{Acc_1, \dots, Acc_K \}$,
for each unseen example $x_{test}$, we generate outputs from the $K$ LLMs denoted by $\{\hat{y}_{test}^1,\dots,\hat{y}_{test}^K\}$ along with the  corresponding instance-specific uncertainty scores.  denoted by $\{u_{test}^1,\dots,u_{test}^K\}$. 
While there can be several fusion strategies, since we are dealing with natural language responses, the simplest one is to example-specific selection from the candidate outputs, i.e., 
\[
\hat{y}_{test} = \hat{y}_{test}^{k^*}, \quad \text{where} \quad {k^*} = \arg\max_{k \in \{1, \dots, K\}} f^k.
\]
Selection criterion $f^k$ could be based on validation set accuracy alone, inverse uncertainty or some combination of both such as $\text{Acc}_k \cdot (1 - u_{test}^k)$  or $\frac{\text{Acc}_k}{u_{test}^k}$.
The first strategy essentially reduces the ensemble to a single most accurate model, while the second one elevates the most confident one. However, both of these approaches are sub-optimal compared to combined criteria, specifically 
\[
f^k = \text{Acc}_k \cdot (1 - u_{test}^k),
\]
which yields the best performance. Experiments with  other combined selection criteria shows similar behavior to the aforementioned one and hence,  we omit the results for brevity.

%There are multiple ways to choose the final answer from these candidate responses - $\hat{y}_{test} = \hat{y}_{test}^j where j = argmax_{k \in \{1,\dots K\}}f^k$. We can define $f^k$ as $Acc_k$ , $1-u_{test}^k$ ,$Acc_k*(1-u_{test}^k)$ or $Acc_k/u_{test}^k$. First strategy essentially reduces the ensemble to a single model having highest accuracy, while second one picks the final answer as the one with the least uncertainty. Both of these are suboptimal compared to  $f_k = Acc_k*(1-u_{test}^k)$ with which we experiment here in this work. Alternative strategy $f_k = Acc_k/u_{test}^k$ shows similar behaviour as above and we omit it's analysis for brevity.

%Although there are multiple ways to aggregate these candidate responses we propose a simple aggregation technique where we choose the final answer as the one with the least uncertainty ie. $\hat{y}_{test} = \hat{y}_{test}^j where j = argmin_{k \in \{1,\dots K\}}u_{test}^k$. Algorithm \ref{alg:selector} presents the pseudo-code of UAF components.

\section{Experiments}
\label{sec:exper}
We investigate the following questions:
\begin{itemize}[leftmargin=*]
    \item \textbf{RQ1:} How effective are uncertainty methods in detecting hallucinations?
    \item \textbf{RQ2:} To what extent do LLMs' relative accuracy and hallucination detection abilities vary across examples?
    \item \textbf{RQ3:} How does UAF perform against SOTA baselines?
    \item \textbf{RQ4:} How does UAF performance vary with ensemble size $K$?
\end{itemize}
\vspace{-4pt}
\subsection{Experimental Setup}
\textbf{Algorithms: }We implement UAF using three uncertainty measures: Perplexity\cite{selgen}, Semantic Entropy\cite{semantic_unc}, and Haloscope\cite{haloscope}, each chosen to represent  distinct category of approaches as mentioned in Section \ref{sec:intro}. We exclude prompting-based methods due to LLMs' tendency to be \textit{overconfident} when verbalizing their confidence \cite{can_llm_unc}. We sample $10\%$ of the data as a validation set for our SELECTOR module and tune the hyperparameter $K$ on this set.
We compare UAF against representation editing methods: ITI \cite{iti} and TruthX \cite{truthx}, as well as contrastive decoding-based methods: DoLa \cite{dola} and SH2 \cite{sh2}. We also evaluate UAF against three prominent ensemble-based methods: Consistent \cite{wang2022self}, MLM \cite{mlm}, and LLM-Blender \cite{llmblender}.We adapt LLM-Blender for zero-shot evaluation by prompting Llama-13B to generate the final answer from the ensemble's candidate responses, instead of training the fusion model.
%\textbf{Uncertainty Algorithms:} We experiment with one method from each category: Perplexity, Semantic Entropy, and Haloscope. We exclude prompting-based methods due to LLMs' tendency to be \textit{overconfident} when verbalizing their confidence \cite{can_llm_unc}.
%\\
%\textbf{Baseline Hallucination mitigation algorithms: }We compare UAF against representation editing methods (ITI \cite{iti}, TruthX \cite{truthx}) and contrastive decoding-based methods (DoLa \cite{dola} and SH2 \cite{sh2}).
%\\
%\textbf{Baseline LLM-ensemble based algorithms:} We evaluate UAF against three prominent zero-shot methods for combining responses from multiple LLMs to address hallucinations: (1) Majority voting to select the most consistent answer (Consistent) \cite{wang2022self}, (2) Semantic similarity to choose the answer most aligned with the question (MLM) \cite{mlm}, and (3) LLM-Blender\cite{llmblender}, which learns to generate the final answer by fusing candidate responses. For zero-shot evaluation, we adapt LLM-Blender by prompting another LLM (Llama-13B) to generate the final answer from all candidate responses.
\\
\textbf{Datasets:} For our evaluation,  we consider two open-book QA datasets,  TruthfulQA\cite{truthfulqa} and FACTOR-news\cite{factor}, which contain 817 and 1,036 multiple-choice QA pairs, respectively, following prior works \cite{iti,dola,truthx} as well as a generative QA dataset TriviaQA \cite{triviaqa} (\textit{rc.nocontext subset}) with $9960$ QA pairs.
%Following \cite{iti,dola,truthx}, to measure truthfulness, we consider two open-book QA datasets for evaluation - TruthfulQA\cite{truthfulqa} and FACTOR-news\cite{factor}. Both of these are multiple choice QA task having $817$ and $1036$ QA pairs respectively. We also evaluate on generative QA dataset TriviaQA\cite{triviaqa} (\textit{rc.nocontext subset}) having $9960$ QA pairs.
\\
\textbf{Metrics:} For multiple-choice tasks (TruthfulQA, FACTOR), correctness is determined based on selection of the gold answer while in case of Trivia-QA, correctness is based on exact match of the ground truth. We report two metrics: \texttt{accuracy}, i.e., the fraction of correctly answered questions, and  \texttt{self assessment of hallucination}, i.e., the area under ROC curve (AUROC) for detecting hallucinations using uncertainty score. 
\\
\textbf{LLM-Pool:} We use open-source LLMs like Llama2-13B, llama 3.2-3B, Alpaca, Vicuna, Mistralv3-7B and Opt-13B. Details of these models can be found in \cite{llm}.

\subsection{RQ1: Effectiveness of uncertainty methods in hallucination detection}
\begin{table}[t]
\centering
\caption{AUROC of uncertainty methods in TruthfulQA. Higher values above $0.5$ is preferred.}
\begin{adjustbox}{width=\columnwidth,center}
\def\arraystretch{1.2}%
\begin{tabular}{l|ccc}
\hline
LLMs         & Perplexity\cite{selgen} & Haloscope\cite{haloscope} & Semantic Entropy\cite{semantic_unc}  \\
\hline
Alpaca-7B    & 0.71      & 0.72        & 0.69                           \\
Vicuna-7B    & 0.8       & 0.87        & 0.82                            \\
Llama2-13B   & 0.68      & 0.71        & 0.72                           \\
Llama3.2-3B  & 0.60    & 0.69         &  0.66                           \\
Mistralv3-7B & 0.51      & 0.78        & 0.63                           \\
Opt-13B      & 0.71      & 0.73        & 0.7                          \\
\hline
\end{tabular}
\end{adjustbox}
\label{table:unc_metrics}
\end{table}

We compare the performance of three uncertainty methods—Perplexity, Semantic Entropy, and Haloscope—across various LLMs in Table \ref{table:unc_metrics}. For each example in TruthfulQA, we prompt the LLMs and mark their responses as either truthful or hallucinatory based on the ground-truth answer. We also generate a corresponding normalized uncertainty score, $\in (0,1)$, for each response using one of the above methods.
We report the area under the receiver operating characteristic curve (AUROC) in Table \ref{table:unc_metrics}, which measures the performance of binary classification of truthful vs. hallucinatory responses by varying the thresholds of uncertainty scores. As shown in Table \ref{table:unc_metrics}, most LLMs achieve high AUROC scores (above the random chance threshold of $0.5$), indicating that uncertainty methods are effective at detecting the models' hallucinatory responses

%\textbf{Different LLMs may be uncertain about different data points.} Columns $>$Vic($\%$) and $>$Mist($\%$) show the percentage of examples where each LLM correctly detects hallucinations missed by Vicuna and Mistralv3-7B, respectively, using the Haloscope uncertainty measure. While Vicuna and Mistralv3 perform well, other LLMs still outperform them in some cases. This highlights that LLMs may not share uncertainty on the same data points, emphasizing the value of pooling uncertainty from multiple models in an ensemble. 

\subsection{RQ2: Variation in accuracy and hallucination detection across LLMs}
We prompt five individual LLMs from our LLM pool on TruthfulQA and compute uncertainty scores using the Haloscope method for each generated response. An incorrect response is considered correctly detected as a hallucination if its uncertainty score is greater than 0.5. In Table \ref{table:variation}, each cell in the $i^{th}$ row and $j^{th}$ column shows a tuple: (1) the percentage of examples where the $j^{th}$ LLM generates the correct response missed by the $i^{th}$ LLM, and (2) the percentage where the $j^{th}$ LLM detects hallucinations missed by the $i^{th}$ LLM. From this, we conclude that each LLM outperforms others in both accuracy and hallucination detection for a substantial proportion of examples. This highlights that the optimal model can vary significantly across data points, emphasizing the value of pooling strengths from multiple models in an ensemble.

\begin{table}[h!]
\centering
\caption{Variation in accuracy and hallucination detection capability across each LLMs in TruthfulQA. }
\begin{adjustbox}{width=\columnwidth,center}
\def\arraystretch{1.2}%
\begin{tabular}{l|ccccc}
\hline
             & Alpaca-7B  & Vicuna-7B  & Llama2-13B  & Llama3.2-3B & Mistralv3-7B \\
             \toprule
Alpaca-7B    & NA         & 6.8 , 17   & 12.3 , 28.2 & 19 , 8.6    & 23.6 , 32.1  \\
Vicuna-7B    & 5.6 , 21.3 & NA         & 14.9 , 33.2 & 17.5 , 19.5 & 21.1 , 16.4  \\
Llama2-13B   & 9.5 , 21   & 8.2 , 19.1 & NA          & 16.6 , 10   & 18 , 25.4    \\
Llama3.2-3B  & 8.6 , 16.6 & 10.2 , 24  & 17 , 28.7   & NA          & 11.5 , 30    \\
Mistralv3-7B & 11 , 19.6  & 6.6 , 20.8 & 15.8 , 25.2 & 8.6 , 14.8  & NA          \\
\bottomrule
\end{tabular}
\end{adjustbox}
\label{table:variation}
\end{table}

\begin{table*}[t!]
\centering
\caption{Results on TruthfulQA, TriviaQA and FACTOR-news datasets. We report accuracy (\%). Best results are bolded.}
\begin{adjustbox}{width=\textwidth,center}

\begin{tabular}{l|cc|cc|ccc|ccc|c}
\hlineB{3}
\multicolumn{1}{c|}{} & \multicolumn{2}{c|}{Representation Editing} & \multicolumn{2}{c|}{Contrastive Decoding} & \multicolumn{3}{c|}{Ensemble} & \multicolumn{3}{c|}{UAF}    & \multicolumn{1}{c}{GPT-4}      \\[4pt]
                     & ITI                & TruthX                & DoLa                & SH2                & Consistent  & MLM  & LLM-Blender & Perplexity & Haloscope & Semantic-Entropy & (as reported \cite{gpt4})\\[3pt]
                    \toprule
TruthfulQA           &    $38.9$                &         $54.2$              &           $33.2$          &          $34.2$          &      $41.7$        &   $29.0$   &   $48.4$     &     $56.5$       &    $\mathbf{60.8}$       &     $51.3$   & $59$ \\[2pt]
TriviaQA             &      $65.9$              &        $69.8$               &         $66.1$            &      $70.2$              &      $65.6$        &   $57.6$   &   $59$     &       $\mathbf{76}$     &     $71.5$      &   $66.4$  & $\mathbf{87}$    \\[2pt]
FACTOR-news               &        $53.3$            &          $65.8$             &        $66.2$             &       $77$             &      $61.4$        &   $50.8$   &    $64.7$    &     $72.5$       &     $\mathbf{81.4}$      &  $69.2$ & -
\\
\hlineB{3}
\end{tabular}
\end{adjustbox}
\label{table:uaf_perf}
\end{table*}

\vspace{-2mm}

\subsection{RQ3: UAF results on benchmark datasets}
Table \ref{table:uaf_perf} shows UAF results across all datasets. Representation editing and contrastive decoding methods are applied to each LLM in our LLM-Pool, reporting the best result, while ensembling uses all LLMs in the pool. We implemented UAF using three different uncertainty measures: Perplexity, Haloscope, and Semantic-Entropy. UAF (with Haloscope) outperforms GPT-4 by $3\%$ on TruthfulQA and surpasses the best baseline by $12\%$. In TriviaQA and Factor-news, UAF outperforms the best baseline by at least $8\%$ and $6\%$, respectively, and narrows the gap with GPT-4. UAF with Semantic-Entropy performs relatively worse, likely due to the method's limited effectiveness in detecting hallucinations. This highlights that weak uncertainty modeling can hurt overall ensemble performance. Interestingly, ensembling strategies like Consistent and LLM-Blender match or outperform some complex hallucination mitigation methods, emphasizing the benefits of ensembling framework.
%\textbf{LLMs have diverse strengths and weakness: } In Table \ref{table:indi}, we report the zero-shot accuracy of each individual LLM on TruthfulQA. Mistralv3-7B and Llama3.2-3B are the top performers. The columns $>Mistralv3(\%)$ and $>Llama3.2(\%)$ show the percentage of examples where other LLMs outperform Mistralv3-7B and Llama3.2-3B, respectively. Despite their strong performance, both still have a substantial number of cases where other LLMs surpass them. This further highlights the benefits of our ensembling framework.
\begin{comment}

\begin{table}[h!]
\centering
\caption{Performance of individual LLMs in TruthfulQA. }
\begin{adjustbox}{width=\columnwidth}

\begin{tabular}{l|cccccc}
\hlineB{3}
            & Alpaca-7B & Vicuna-7B & Llama2-13B & Llama3.2-3B & Mistralv3-7B & Opt-13B \\[4pt]
\toprule
Accuracy $\uparrow$ &   $27.8$    &     $33.4$    &    $34.6$       &   $41.0$    &   $48.7$   &      $35.1$  \\[2pt]
$>Mistralv3(\%)$ $\uparrow$ & $11$     & $16.6$        & $15.8$          & $8.6$         &   NA    & $12.8$  \\[2pt]
$>Llama3.2(\%)$ $\uparrow$ & $8.6$     & $10.2$        & $17.0$          & NA        &   $11.5$    & $12.2$  \\[2pt]
\bottomrule
\end{tabular}
\end{adjustbox}
\label{table:indi}
\end{table}

\vspace{-2mm}
\end{comment}

\subsection{RQ4: Ablation study with hyperparameter $K$}
\begin{figure}[h!]
\centering
\includegraphics[width=.3\textwidth]{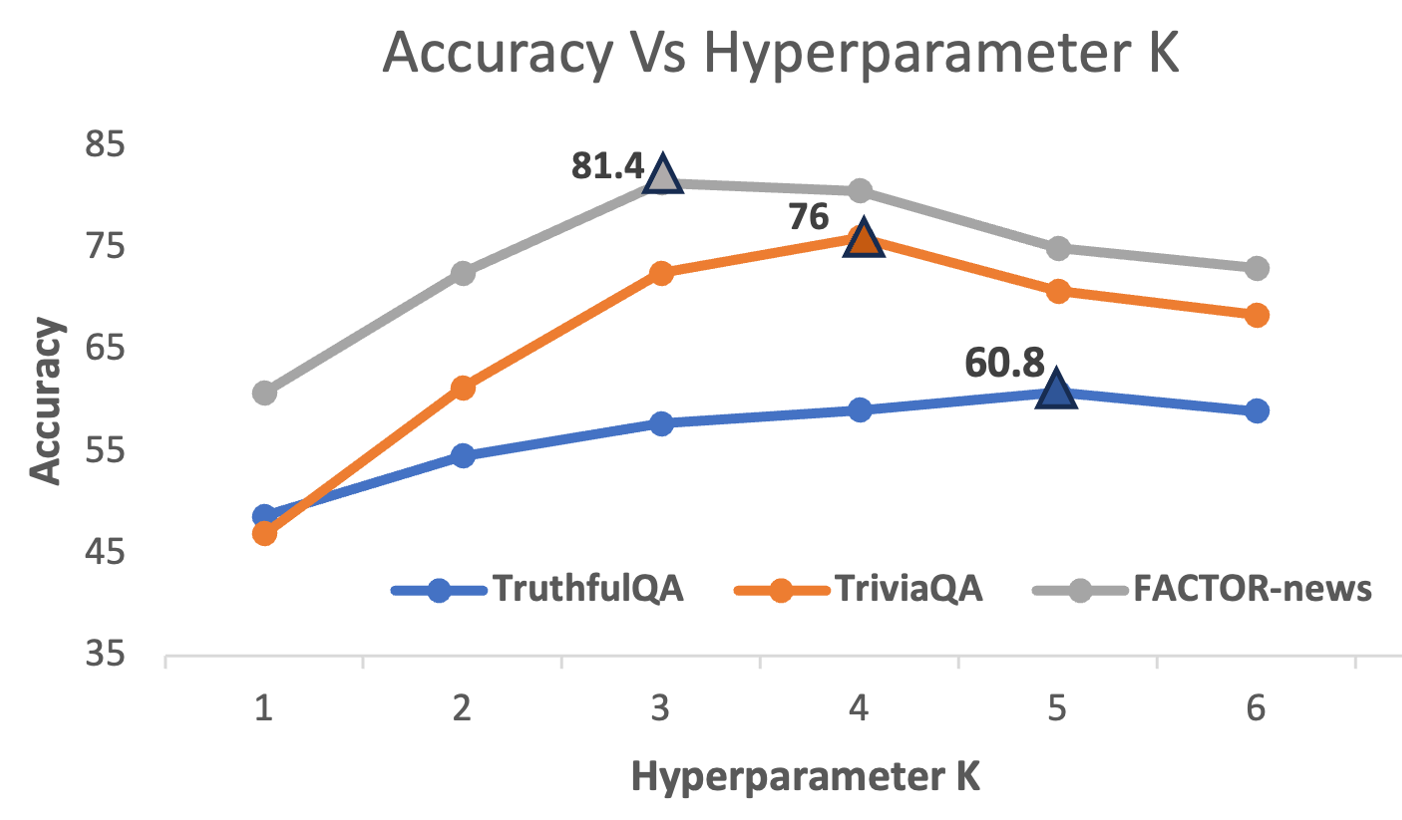}
\caption{Impact of $K$ on UAF performance.}
\label{fig:hypk}
\end{figure}

Figure \ref{fig:hypk} shows the impact of varying $K$, the number of LLMs chosen by SELECTOR, on UAF accuracy across  datasets. For TruthfulQA, the highest accuracy ($60.8\%$) is achieved with top 5 LLMs (Mistralv3-7B, Vicuna, Llama3.2-3B, Opt-13B, and Llama2-13B). 
For TriviaQA and FACTOR, the top 4 LLMs (Llama2-13B, Mistralv3-7B, Alpaca, Llama3.2-3B) and top 3 LLMs (Mistralv3-7B, Llama3.2-3B, Llama2-13B) yield optimal results, with performance sharply declining for larget $K$. These results highlight the importance of the SELECTOR module in the UAF method, demonstrating that carefully choosing a small subset of high-performing LLMs enhances ensemble performance more effectively than using the entire LLM pool.

\vspace{-2mm}

\section{Conclusion}
In this work, we introduced an ensemble framework UAF that reduces hallucinations in factoid question answering (QA) by leveraging both the accuracy and self-assessment capabilities of multiple LLMs. By incorporating uncertainty estimation, UAF strategically combines model responses, improving factual accuracy on multiple benchmark datasets. Since UAF combines the responses of multiple LLMs, the inference time scales linearly with the number of models in the ensemble, requiring multiple forward passes through each model to generate responses and uncertainty estimates. Additionally, the uncertainty estimation itself may introduce some overhead, depending on the specific method used to compute it. Despite the added computational complexity of ensemble models, the significant improvement in factual accuracy justifies the approach. Future work could explore dynamic LLM selection based on each data point and integrate reinforcement learning for better adaptation to diverse data, enhancing both accuracy and efficiency.

%In this work, the computational complexity primarily arises from the ensemble-based nature of (UAF) framework. Since UAF combines the responses of multiple LLMs, the inference time scales linearly with the number of models in the ensemble, requiring multiple forward passes through each model to generate responses and uncertainty estimates. Additionally, the uncertainty estimation itself may introduce some overhead, depending on the specific method used to compute it. Despite these increased computational demands, UAF's design strives to balance efficiency with performance gains, showing significant improvements in factual accuracy without overwhelming the system's resources. However, in practical deployments, especially with larger ensembles, it is important to account for the trade-off between the additional computational cost and the accuracy gains achieved through more reliable and moderated predictions.
%Future work could explore dynamic LLM selection based on each data point and integrate reinforcement learning for better adaptation to diverse data, enhancing both accuracy and efficiency.

%\bibliographystyle{abbrv}
%\bibliography{sample-base}
%\vspace{1cm}
%\input{appendix}
\bibliographystyle{ACM-Reference-Format}
\balance
\bibliography{main}


\end{document}